\documentclass[11pt,a4paper]{article}
\usepackage[OT1]{fontenc}
\usepackage[hyperref]{naaclhlt2018}
\usepackage{times}
\usepackage{amsmath}
\usepackage{latexsym}
\usepackage{graphicx}

\usepackage{url}
\usepackage{enumitem}
\usepackage[hang,flushmargin]{footmisc}

\setlist[itemize]{leftmargin=*}

\aclfinalcopy % Uncomment this line for all SemEval submissions

\title{Yuanfudao at SemEval-2018 Task 11: Three-way Attention and Relational Knowledge for Commonsense Machine Comprehension}

\author{Liang Wang \qquad Meng Sun \qquad Wei Zhao \qquad Kewei Shen \qquad Jingming Liu \\
    Yuanfudao Research \\
    Beijing, China \\
    \{wangliang01,sunmeng,zhaowei01,shenkw,liujm\}@fenbi.com\\}

\date{}

\begin{document}
\maketitle
\begin{abstract}
This paper describes our system for
\emph{SemEval-2018 Task 11: Machine Comprehension using Commonsense Knowledge} ~\cite{SemEval2018Task11}.
We use \textbf{Three}-way \textbf{A}ttentive \textbf{N}etworks (\emph{TriAN})
to model interactions between the passage, question and answers.
To incorporate commonsense knowledge,
we augment the input with relation embedding
from the graph of general knowledge \emph{ConceptNet} ~\cite{speer2017conceptnet}.
As a result,
our system achieves state-of-the-art performance with $83.95\%$ accuracy on the official test data.
Code is publicly available at \url{https://github.com/intfloat/commonsense-rc}.
\end{abstract}

\section{Introduction}

It is well known that humans have a vast amount of commonsense knowledge acquired from everyday life.
For machine reading comprehension,
natural language inference and many other NLP tasks,
commonsense reasoning is one of the major obstacles to make machines as intelligent as humans.

A large portion of previous work focus on commonsense knowledge acquisition
with unsupervised learning ~\cite{chambers2008unsupervised,tandon2017webchild}
or crowdsourcing approach ~\cite{singh2002open,wanzare2016descript}.
\emph{ConceptNet} ~\cite{speer2017conceptnet}, \emph{WebChild} ~\cite{tandon2017webchild}
and \emph{DeScript} ~\cite{wanzare2016descript} etc are all publicly available knowledge resources.
However,
resources based on unsupervised learning tend to be noisy,
while crowdsourcing approach has scalability issues.
There is also some research on incorporating knowledge into NLP tasks
such as reading comprehension ~\cite{lin2017reasoning,yang2017leveraging}
neural machine translation ~\cite{zhang2017prior} and text classification ~\cite{zhang2017exploiting} etc.
Though experiments show performance gains over baselines,
these gains are often quite marginal over the state-of-the-art system without external knowledge.

In this paper,
we present \textbf{Three}-way \textbf{A}ttentive \textbf{N}etworks(\emph{TriAN})
for multiple-choice commonsense reading comprehension.
The given task requires modeling interactions between the passage,
question and answers.
Different questions need to focus on different parts of the passage,
attention mechanism is a natural choice
and turns out to be effective for reading comprehension.
Due to the relatively small size of training data,
\emph{TriAN} use word-level attention
and consists of only one layer of LSTM ~\cite{hochreiter1997long}.
Deeper models result in serious overfitting and poor generalization empirically.

To explicitly model commonsense knowledge,
relation embeddings based on ConceptNet~\cite{speer2017conceptnet} are used as additional features.
ConceptNet is a large-scale graph of general knowledge
from both crowdsourced resources and expert-created resources.
It consists of over $21$ million edges and $8$ million nodes.
ConceptNet shows state-of-the-art performance on tasks like word analogy and word relatedness.

Besides,
we also find that pretraining our network on other datasets helps to improve the overall performance.
There are some existing multiple-choice English reading comprehension datasets
contributed by NLP community such as \emph{MCTest} ~\cite{richardson2013mctest} and \emph{RACE} ~\cite{lai2017race}.
Although those datasets don't focus specifically on commonsense comprehension,
they provide a convenient way for data augmentation.
Augmented data can be used to learn shared regularities of reading comprehension tasks.

Combining all of the aforementioned techniques,
our system achieves competitive performance on the official test set.

\begin{figure*}[ht]
\begin{center}
 \includegraphics[width=.99\linewidth]{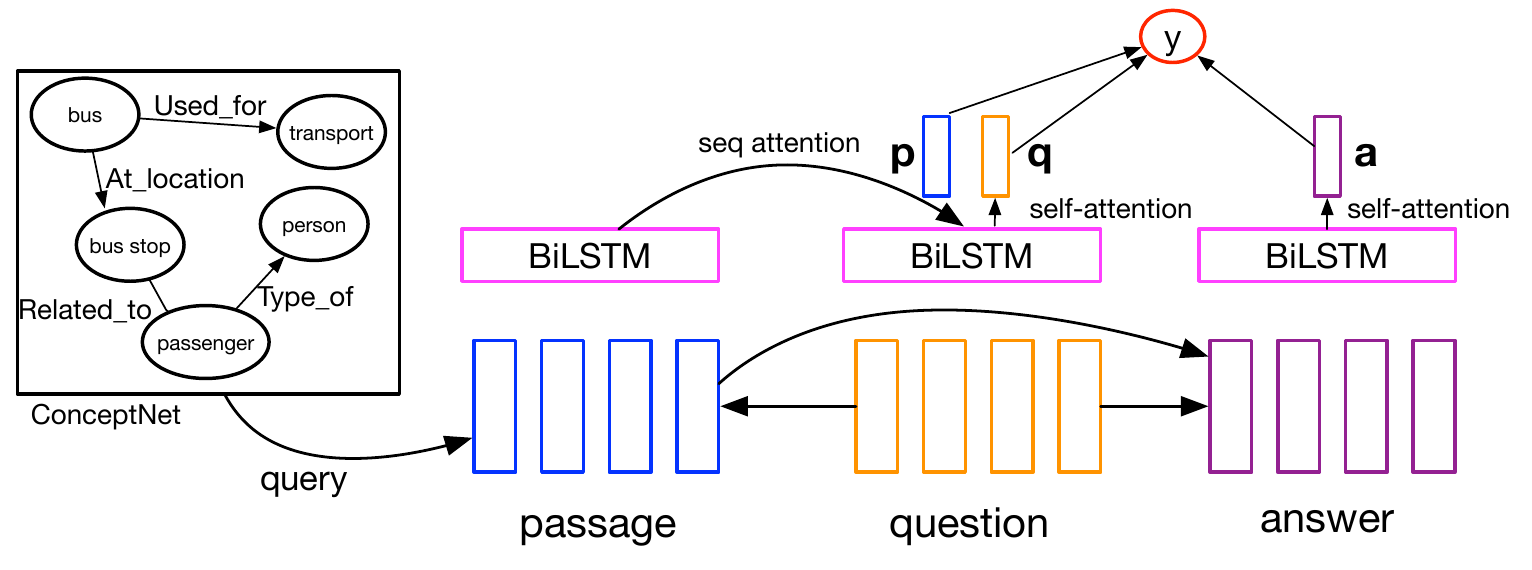}
 \caption{\emph{TriAN} Model Architecture.}
 \label{fig:TriAN}
\end{center}
\end{figure*}

\section{Model}
The overall architecture of \emph{TriAN} is shown in Figure ~\ref{fig:TriAN}.
It consists of an input layer,
an attention layer and an output layer.

\noindent
\textbf{Input Layer. }
A training example consists of a passage $\{P_i\}_{i=1}^{|P|}$,
a question $\{Q_i\}_{i=1}^{|Q|}$,
an answer $\{A_i\}_{i=1}^{|A|}$ and a label $y^* \in \{0, 1\}$ as input.
$P$, $Q$ and $A$ are all sequences of word indices.
For a word $P_i$ in the given passage,
the input representation of $P_i$ is the concatenation of several vectors:
\begin{itemize}
\setlength\itemsep{0em}
\item \textbf{GloVe embeddings.}
Pretrained $300$-dimensional \emph{GloVe} vector $\mathbf{E}_{P_i}^{glove}$.
\item \textbf{Part-of-speech and named-entity embeddings.}
Randomly initialized $12$-dimensional part-of-speech embedding $\mathbf{E}_{P_i}^{pos}$
and $8$-dimensional named-entity embedding $\mathbf{E}_{P_i}^{ner}$.
\item \textbf{Relation embeddings.}
Randomly initialized $10$-dimensional relation embedding $\mathbf{E}_{P_i}^{rel}$.
The relation is determined by querying ConceptNet
whether there is an edge between $P_i$ and
any word in question $\{Q_i\}_{i=1}^{|Q|}$ or answer $\{A_i\}_{i=1}^{|A|}$.
If there exist multiple different relations,
just randomly choose one.
\item \textbf{Handcrafted features.}
We also add logarithmic term frequency feature and co-occurrence feature $\mathbf{f}_{P_i}$.
Term frequency is calculated based on English Wikipedia.
Co-occurrence feature is a binary feature which is true
if $P_i$ appears in question $\{Q_i\}_{i=1}^{|Q|}$ or answer $\{A_i\}_{i=1}^{|A|}$.
\end{itemize}
The input representation for $P_i$ is $\mathbf{w}_{P_i}$:
\begin{equation}
\mathbf{w}_{P_i} = [\mathbf{E}_{P_i}^{glove}; \mathbf{E}_{P_i}^{pos}; \mathbf{E}_{P_i}^{ner}; \mathbf{E}_{P_i}^{rel}; \mathbf{f}_{P_i}]
\end{equation}
In a similar way,
we can get input representation for question $\mathbf{w}_{Q_i}$ and answer $\mathbf{w}_{A_i}$.

\noindent
\textbf{Attention Layer. }
We use word-level attention to model interactions between the given passage $\{P_i\}_{i=1}^{|P|}$,
the question $\{Q_i\}_{i=1}^{|Q|}$ and the answer $\{A_i\}_{i=1}^{|A|}$.
First,
let's define a sequence attention function ~\cite{chen2017reading}:
\begin{equation}
\begin{aligned}
Att_{seq}(\mathbf{u},\ \{\mathbf{v}_i\}_{i=1}^{n}) = \sum_{i=1}^{n}&{\alpha_i \mathbf{v}_i} \\
\alpha_i = \mathrm{softmax}_i(f(\mathbf{W}_1\mathbf{u})^T & f(\mathbf{W}_1\mathbf{v}_i))
\end{aligned}
\end{equation}

$\mathbf{u}$ and $\mathbf{v}_i$ are vectors
and $\mathbf{W}_1$ is a matrix.
$f$ is a non-linear activation function
and is set to $ReLU$.

Question-aware passage representation $\{\mathbf{w}_{P_i}^q\}_{i=1}^{|P|}$ can be calculated as:
$\mathbf{w}_{P_i}^q = Att_{seq}(\mathbf{E}_{P_i}^{glove}, \{\mathbf{E}_{Q_i}^{glove}\}_{i=1}^{|Q|})$.
Similarly,
we can get passage-aware answer representation $\{\mathbf{w}_{A_i}^p\}_{i=1}^{|A|}$
and question-aware answer representation $\{\mathbf{w}_{A_i}^q\}_{i=1}^{|A|}$.
Then three BiLSTMs are applied
to the concatenation of those vectors to model the temporal dependency:
\begin{equation}
\begin{aligned}
\mathbf{h}^q & = \mathrm{BiLSTM}(\{\mathbf{w}_{Q_i}\}_{i=1}^{|Q|}) \\
\mathbf{h}^p & = \mathrm{BiLSTM}(\{[\mathbf{w}_{P_i}; \mathbf{w}_{P_i}^q]\}_{i=1}^{|P|}) \\
\mathbf{h}^a & = \mathrm{BiLSTM}(\{[\mathbf{w}_{A_i}; \mathbf{w}_{A_i}^p; \mathbf{w}_{A_i}^q]\}_{i=1}^{|A|}) \\
\end{aligned}
\end{equation}

$\mathbf{h}^p, \mathbf{h}^q, \mathbf{h}^a$ are the new representation vectors
that incorporates more context information.

\noindent
\textbf{Output Layer. }
Question sequence and answer sequence representation $\mathbf{h}^q, \mathbf{h}^a$
are summarized into fixed-length vectors with self-attention ~\cite{yang2016hierarchical}.
Self-attention function is defined as follows:
\begin{equation}
\begin{aligned}
Att_{self}(\{\mathbf{u}_i\}_{i=1}^n) = \sum_{i=1}^{n}&{\alpha_i \mathbf{u}_i} \\
\alpha_i = \mathrm{softmax}_i(\mathbf{W}_2^T \mathbf{u}_i)&
\end{aligned}
\end{equation}

Then we have question representation $\mathbf{q} = Att_{self}(\{\mathbf{h}_i^q\}_{i=1}^{|Q|})$,
answer representation $\mathbf{a} = Att_{self}(\{\mathbf{h}_i^a\}_{i=1}^{|A|})$
and passage representation $\mathbf{p} = Att_{seq}(\mathbf{q}, \{\mathbf{h}_i^p\}_{i=1}^{|P|})$.
The final output $y$ is based on their bilinear interactions:
\begin{equation}
y = \sigma(\mathbf{p}^T\mathbf{W}_3\mathbf{a} + \mathbf{q}^T\mathbf{W}_4\mathbf{a})
\end{equation}

\noindent
\textbf{Model Learning. }
We first pretrain \emph{TriAN} on \emph{RACE} dataset for $10$ epochs.
Then our model is fine-tuned on official training data.
Standard cross entropy function is used as the loss function to minimize.

\section{Experiments}
\subsection{Setup}
\noindent
\textbf{Data. }
For data preprocessing,
we use \emph{spaCy}~\footnote{\url{https://github.com/explosion/spaCy}}
for tokenization, part-of-speech tagging and named-entity recognition.
Statistics for official dataset \emph{MCScript} ~\cite{MCScript} are shown in Table ~\ref{table:dataset}.
\emph{RACE}~\footnote{\url{http://www.cs.cmu.edu/~glai1/data/race/}}
dataset is used for network pretraining.
English stop words are ignored when computing handcrafted features.
Input word embeddings are initialized with $300$-dimensional
\emph{GloVe} ~\cite{pennington2014glove} vectors ~\footnote{\url{http://nlp.stanford.edu/data/glove.840B.300d.zip}}.

\begin{table}[ht]
\centering
\begin{tabular}{c|ccc}
 \hline
               & train & dev  & test \\ \hline
\# of examples & 9731  & 1411 & 2797 \\ \hline
\end{tabular}
\caption{Official dataset statistics.}
\label{table:dataset}
\end{table}

\noindent
\textbf{Hyperparameters. }
Our model \emph{TriAN} is implemented based on \emph{PyTorch} ~\footnote{\url{http://pytorch.org/}}.
Models are trained on a single GPU(Tesla P40)
and each epoch takes about $80$ seconds.
Only the word embeddings of top $10$ frequent words are fine-tuned during training.
The dimension of both forward and backward LSTM hidden state is set to $96$.
Dropout rate is set to $0.4$ for both input embeddings and BiLSTM outputs ~\cite{srivastava2014dropout}.
For parameter optimization,
we use \emph{Adamax}~\cite{kingma2014adam} with an initial learning rate $2\times10^{-3}$.
Learning rate is then halved after $10$ and $15$ training epochs.
The model converges after $50$ epochs.
Gradients are clipped to have a maximum L2 norm of $10$.
Minibatch with batch size $32$ is used.
Hyperparameters are optimized by random search strategy~\cite{bergstra2012random}.
Our model is quite robust over a wide range of hyperparameter configurations.

\subsection{Main Results}

The experimental results are shown in Table ~\ref{table:main_results}.
Human performance is shared by task organizers.
For \emph{TriAN-ensemble},
we average the output probabilities of 9 models
trained with the same datasets and network architecture
but different random seeds.
\emph{TriAN-ensemble} is the model
that we used for official submission.

\begin{table}[ht]
\centering
\begin{tabular}{l|ll}
 \hline
\multicolumn{1}{c|}{model}     & \multicolumn{1}{c}{dev} & \multicolumn{1}{c}{test} \\ \hline
Random   & 50.00\%   & 50.00\%    \\ \hline
TriAN-RACE   & 64.78\%   & 64.28\%  \\ \hline
TriAN-single   & 83.84\%   & 81.94\%    \\ \hline
TriAN-ensemble & \textbf{85.27\%}   & \textbf{83.95\%}    \\ \hline \hline
HFL        & \multicolumn{1}{c}{--}   & 84.13\%    \\ \hline
Human         &  \multicolumn{1}{c}{--}    &  98.00\%  \\ \hline
\end{tabular}
\caption{Main results.
\emph{TriAN-RACE} only use \emph{RACE} dataset for training;
\emph{HFL} is the 1st place team for \emph{SemEval-2018 Task 11}.
The evaluation metric is accuracy.}
\label{table:main_results}
\end{table}

From Table \ref{table:main_results},
we can see that even though \emph{RACE} dataset contains nearly $100k$ questions,
\emph{TriAN-RACE} achieves quite poor results.
The accuracy on development set is only $64.78\%$,
which is worse than most participants' systems.
However,
pretraining acts as a way of implicit knowledge transfer
and is beneficial for overall performance,
as will be seen in Section ~\ref{section:ablation}.
The accuracy of our system \emph{TriAN-ensemble}
is very close to the 1st place team \emph{HFL} with $0.18\%$ difference.
Yet there is still a large gap between machine learning models and human.

We also compared the performances of shallow and deep TriAN models.
On datasets such as \emph{SQuAD} ~\cite{rajpurkar2016squad},
deep models typically works better than shallow ones.
Notice that the attention layer in our proposed \emph{TriAN} model
can be stacked multiple times
if we treat the output vectors of BiLSTMs as new input representations.

\begin{table}[ht]
\centering
\begin{tabular}{l|ll}
 \hline
\multicolumn{1}{c|}{model} & \multicolumn{1}{c}{dev} & \multicolumn{1}{c}{test} \\ \hline
1-layer TriAN-single       & \textbf{83.84\%}  &  \textbf{81.94\%}  \\ \hline
2-layer TriAN-single       & 82.71\%    & 80.55\%  \\ \hline
\end{tabular}
\caption{Accuracy comparison of shallow and deep TriAN models.}
\label{table:layers}
\end{table}

Maybe a little bit surprising,
Table ~\ref{table:layers} shows that
\emph{2-layer TriAN} model performs worse than \emph{1-layer TriAN}.
One possible explanation is that
the labeled dataset is relatively small
and deeper models tend to easily overfit.

\subsection{Ablation Study} \label{section:ablation}

The input representation consists of several components:
part-of-speech embedding,
relation embedding
and handcrafted features etc.
We conduct an ablation study to investigate the effects of each component.
The results are in Table ~\ref{table:input_ablation}.

\begin{table}[ht]
\centering
\begin{tabular}{l|ll}
 \hline
\multicolumn{1}{c|}{model}     & \multicolumn{1}{c}{dev} & \multicolumn{1}{c}{test} \\ \hline
TriAN-single    & \textbf{83.84\%}   & \textbf{81.94\%}  \\ \hline \hline
w/o pretraining & 82.71\%   & 80.51\%  \\ \hline
w/o ConceptNet  & 82.78\%   & 81.08\%    \\ \hline
w/o POS         & 82.84\%   & 81.27\%    \\ \hline
w/o features    & 82.92\%   & 81.35\% \\ \hline
w/o NER         & 83.60\%   & 81.87\%    \\ \hline
\end{tabular}
\caption{Ablation study for input representation.}
\label{table:input_ablation}
\end{table}

Pretraining on \emph{RACE} dataset turns out to be the most important factor.
Without pretraining,
the accuracy drops by more than $1\%$ on both development and test set.
Relation embeddings based on ConceptNet make approximately $1\%$ difference.
Part-of-speech and named-entity embeddings are also helpful.
In fact,
combining input representations from multiple sources
has been a standard practice for reading comprehension tasks.

At attention layer,
our proposed \emph{TriAN} involves applying several attention functions
to model interactions between different text sequences.
It would be interesting to examine the importance of each attention function,
as shown in Table ~\ref{table:attention_ablation}.

\begin{table}[ht]
\hskip-0.3cm
\centering
\scalebox{0.9}{
\begin{tabular}{l|ll}
 \hline
\multicolumn{1}{c|}{model}     & \multicolumn{1}{c}{dev} & \multicolumn{1}{c}{test} \\ \hline
TriAN-single                   & \textbf{83.84\%}   & 81.94\% \\ \hline \hline
w/o passage-question attention & 83.51\%   & \textbf{82.20\%}    \\ \hline
w/o passage-answer attention   & 83.07\%   & 81.39\%    \\ \hline
w/o question-answer attention  & 83.23\%   & 81.84\%    \\ \hline
w/o attention                  & 81.93\%   & 80.65\%    \\ \hline
\end{tabular}}
\caption{Ablation study for attention.
The last one ``\emph{w/o attention}'' removes all word-level attentions.}
\label{table:attention_ablation}
\end{table}

Interestingly,
removing any of the three word-level sequence attentions does not seem to hurt the performance much.
In fact,
removing passage-question attention even results in higher accuracy on test set than \emph{TriAN-single}.
However,
if we remove all word-level attentions,
the performance drastically drops by $1.9\%$ on development set and $1.3\%$ on test set.

\subsection{Discussion}

Even though our system is built for commonsense reading comprehension,
it doesn't have any explicit knowledge reasoning component.
Relation embeddings based on ConceptNet only serve as additional input features.
Methods like event calculus ~\cite{mueller2014commonsense} are more rigorous mathematically
and resemble the way of how human brain works.
The problem of event calculus is that
it requires large amounts of domain-specific axioms
and therefore doesn't scale well.

Another limitation is that
our system relies on hard-coded commonsense knowledge bases,
just like most systems for commonsense reasoning.
For humans,
commonsense knowledge comes from constant interactions with the real-world environment.
From our point of view,
it is quite hopeless to enumerate all of them.

There are a lot of reading comprehension datasets available.
When the size of training data is relatively small like this \emph{SemEval-2018 task},
transfer learning among different datasets is a useful technique.
This paper shows that pretraining is a simple and effective method.
However,
it still remains to be seen whether there is a better alternative approach.

\section{Conclusion}
In this paper,
we present the core ideas and design philosophy
for our system \emph{TriAN} at \emph{SemEval-2018 Task 11: Machine Comprehension using Commonsense Knowledge}.
We build upon recent progress on neural models for reading comprehension
and incorporate commonsense knowledge from ConceptNet.
Pretraining and handcrafted features are also proved to be helpful.
As a result,
our proposed model \emph{TriAN} achieves near state-of-the-art performance.

\section*{Acknowledgements}
We would like to thank SemEval 2018 task organizers
and several anonymous reviewers for their helpful comments.

\bibliography{semeval2018}
\bibliographystyle{acl_natbib}

\end{document}